\documentclass[10pt,twocolumn,letterpaper]{article}

\usepackage{silence}

\usepackage{cvpr}
\usepackage{times}
\usepackage{epsfig}
\usepackage{graphicx}
\usepackage{amsmath}
\usepackage{amssymb}
\usepackage{mystyle}

\usepackage[table]{xcolor}
\usepackage{balance}
\usepackage{tabularx}
\newcolumntype{Y}{>{\centering\arraybackslash}X}

\usepackage[pagebackref=true,breaklinks=true,letterpaper=true,colorlinks,bookmarks=false]{hyperref}

\cvprfinalcopy 

\def\cvprPaperID{9} 

\ifcvprfinal\fi
\begin{document}

\title{Tensor Contraction Layers for Parsimonious Deep Nets
}

\author{Jean Kossaifi\\
Amazon AI\\
Imperial College London\\
{\tt\small jean.kossaifi@imperial.ac.uk}
\and
Aran Khanna\\
Amazon AI\\
{\tt\small arankhan@amazon.com}
\and
Zachary C. Lipton\\
Amazon AI\\
University of California, San Diego\\
{\tt\small zlipton@cs.ucsd.edu}
\and
Tommaso Furlanello\\
Amazon AI\\
University of Southern California\\
{\tt\small furlanel@usc.edu}
\and
Anima Anandkumar\\
Amazon AI\\
California Institute of Technology\\
{\tt\small anima@amazon.com}
}

\maketitle
\thispagestyle{empty}

\begin{abstract}
Tensors offer a natural representation 
for many kinds of data frequently encountered in machine learning. 
Images, for example, are naturally represented as third order tensors, 
where the modes correspond to height, width, and channels. 
Tensor methods are noted 
for their ability to discover multi-dimensional dependencies, 
and tensor decompositions in particular, have been used to produce 
compact low-rank approximations of data. 
In this paper, we explore the use of tensor contractions 
as neural network layers and investigate several ways 
to apply them to activation tensors. 
Specifically, we propose the Tensor Contraction Layer (TCL), 
the first attempt to incorporate tensor contractions 
as end-to-end trainable neural network layers. 
Applied to existing networks, 
TCLs reduce the dimensionality of the activation tensors 
and thus the number of model parameters. 
We evaluate the TCL on the task of image recognition, 
augmenting two popular networks (AlexNet, VGG). 
The resulting models are trainable end-to-end. 
Applying the TCL to the task of image recognition,
using the CIFAR100 and ImageNet datasets,
we evaluate the effect of parameter reduction
via tensor contraction on performance.
We demonstrate significant model compression 
without significant impact on the accuracy 
and, in some cases, improved \mbox{performance}.
\end{abstract}

\section{Introduction}
Following their successful application to computer vision, 
speech recognition, and natural language processing, 
deep neural networks have become ubiquitous 
in the machine learning community.
And yet many questions remain unanswered:
Why do deep neural networks work?
How many parameters are really necessary 
to achieve state of the art performance?

Recently, tensor methods have been used 
in attempts to better understand 
the success of deep neural networks 
\cite{cohen2015expressive,haeffele2015global}.
One class of broadly useful techniques within tensor methods
are tensor decompositions.
While the properties of tensors have long been studied,
in the past decade they have come to prominence 
in machine learning in such varied applications 
as learning latent variable models \cite{anandkumar2014tensor}, 
and developing recommender systems \cite{karatzoglou2010multiverse}.
Several recent papers apply tensor learning 
and tensor decomposition to deep neural networks 
for the purpose of devising neural network learning algorithms
with theoretical guarantees of convergence 
\cite{sedghi2016training,janzamin2015generalization}.

Other lines of research have investigated 
practical applications of tensor decomposition
to deep neural networks with aims including
multi-task learning \cite{yang2016deep}, 
sharing residual units \cite{chen2017sharing},
and speeding up convolutional neural networks \cite{lebedev2014speeding}. 
Several recent papers apply decompositions 
for either initialization \cite{yang2016deep} 
or post-training \cite{novikov2015tensorizing}.
These techniques then often require additional fine-tuning 
to compensate for the loss of information \cite{yong2015compression}.
However, to our knowledge, 
no attempt has been made to apply tensor contractions 
as a generic layer directly on the activations or weights 
of a deep neural network and to train the resulting network end-to-end. 

In deep convolutional neural networks, 
the output of each layer is a tensor. 
We posit that tensor algebraic techniques
can 
exploit multidimensional dependencies 
in the activation tensors. 
We propose to leverage that structure 
by incorporating Tensor Contraction Layers (TCLs) into neural networks. 
Specifically, in our experiments, 
we apply TCLs directly to the third-order activation tensors
produced by the final convolutional layer of an image recognition network.
Traditional networks flatten this activation tensor,
passing it to subsequent fully-connected layers.
However, the flattening process loses information 
about the multidimensional structure of the tensor. 
Our experiments show that incorporating TCLs 
into several popular deep convolutional networks 
can improve their performance, 
despite reducing the number of parameters. 
Moreover, inference on TCL-equipped networks, 
which contain less parameters, 
requires considerably fewer floating point operations.

We organize the rest of this paper as follows: 
Section~\ref{seq:math} introduces prerequisite concepts 
needed to understand the TCL; 
Section~\ref{seq:TCL} explains the TCL in detail;  Section~\ref{seq:experiments} experimentally evaluates the TCL.

\subsection{Tensor Contraction}
\label{seq:math}

\paragraph{Notation: }
We define tensors as multidimensional arrays, 
denoting first-order tensors \(\myvector{v}\) as \emph{vectors}, 
second-order tensors \(\mymatrix{M}\) as \emph{matrices} 
and by \(\mytensor{X}\), 
refer to tensors of order 3 or greater. 
\(\mymatrix{M}\myT\) denotes the transpose of \(\mymatrix{M}\).

\paragraph{Tensor unfolding: }
Given a tensor,
\( \mytensor{X} \in \myR^{D_1 \times D_2 \times \cdots \times D_N}\), 
the mode-\(n\) unfolding of \(\mytensor{X}\) is a matrix \(\mymatrix{X}_{[n]} \in \myR^{D_n, D_{(-n)}}\), 
with \(D_{(-n)} = \prod_{\substack{k=1,\\k \neq n}}^N D_k\)
and is defined by the mapping from element
\( (d_1, d_2, \cdots, d_N)\) to \((d_n, e)\), with 
\(
e = \sum_{\substack{k=1,\\k \neq n}}^N d_k \times \prod_{m=k+1}^N D_m 
\).

\paragraph{n-mode product: }
For a tensor \(\mytensor{X} \in \myR^{D_1 \times D_2 \times \cdots \times D_N}\) and a matrix \( \mymatrix{M} \in \myR^{R \times D_n} \), the n-mode product of  \(\mytensor{X}\) by \( \mymatrix{M}\) is a tensor of size \(\left(D_1 \times \cdots \times D_{n-1} \times R \times D_{n+1} \times \cdots \times D_N\right)\) and
can be expressed using the unfolding of \(\mytensor{X}\) and the classical matrix multiplication as:
\begin{equation}
	\mytensor{X} \times_n \mymatrix{M} = \mymatrix{M} \mytensor{X}_{[n]} \in \myR^{D_1 \times \cdots \times D_{n-1} \times R \times D_{n+1} \times \cdots \times D_N}
\end{equation}

\paragraph{Tensor contraction: }
Given a tensor \(\mytensor{X} \in \myR^{D_1 \times D_2 \times \cdots \times D_N} \), 
we can decompose it into a low-dimensional core tensor \(\mytensor{G} \in \myR^{R_1 \times R_2 \times \cdots \times R_N}\) through projection along each of its modes by projection factors 
\( \left( \mymatrix{U}^{(1)}, \cdots,\mymatrix{U}^{(N)} \right) \), with \(\mymatrix{U}^{(k)} \in \myR^{R_k, D_k}, k \in (1, \cdots, N)\).
In other words, we can write:
\begin{equation}
\mytensor{G} = 
\mytensor{X} \times_1 \mymatrix{U}^{(1)} 
		  \times_2  \mymatrix{U}^{(2)} \times
		  \cdots
          \times_N \mymatrix{U}^{(N)}
\end{equation}
or, in short:
\begin{equation}
\mytensor{G} = 
\mytucker{\mytensor{X}}{\mymatrix{U}^{(1)},
		  \cdots,
          \mymatrix{U}^{(N)}}
\end{equation}

In the case of tensor decomposition, 
the factors of the contraction 
are obtained by solving a least squares problem. 
In particular, closed form solutions 
can be obtained for the factor by considering the \(n-\)mode unfolding of \(\mytensor{X}\) 
that can be expressed as:
\footnotesize
\begin{equation}
  \mymatrix{G}_{[n]} = \mymatrix{U}^{(n)} \mymatrix{X}_{[n]}
    				   \left(\mymatrix{U}^{(1)}
                       \otimes \cdots 
                       \mymatrix{U}^{(n-1)} 
                       \otimes \mymatrix{U}^{(n+1)}
                       \otimes \cdots
                       \otimes \mymatrix{U}^{(N)} \right)^T
\label{eq:unfold_tucker}
\end{equation}
\normalsize

We refer the interested reader 
to the seminal work of Kolda and Bader \cite{kolda2009tensor}.

\subsection{Networks with Large fully connected layers}
Many popular convolutional neural networks for computer vision, 
e.g. AlexNet, ResNet, and Inception, 
require hundreds of millions of parameters 
to achieve the reported results. 
This can be problematic
when running these networks for inference 
on resource-constrained devices, 
where it may not be easy to execute 
hundreds of millions of calculations 
just to classify a single image. 

While these widely used architectures 
exhibit considerable variety, 
they also exhibit some commonalities.
Often, they consist of blocks containing 
convolution, activation and pooling layers 
followed by fully-connected layers 
before the final classification layer. 
Both the popular networks AlexNet \cite{alexnet} 
and VGG \cite{vgg} follow this meta-architecture, 
with both containing two fully-connected layers 
of $4096$ hidden units each. 
In both networks, these fully-connected layers 
hold over $80$ percent of the parameters. 
In VGG, the hidden units contain 119,545,856
of the 138,357,544 total parameters, 
and in AlexNet the hidden units contain 54,534,144 
out the 62,378,344 total parameters.

\begin{figure}
\begin{center}
	\includegraphics[width=8.3cm]{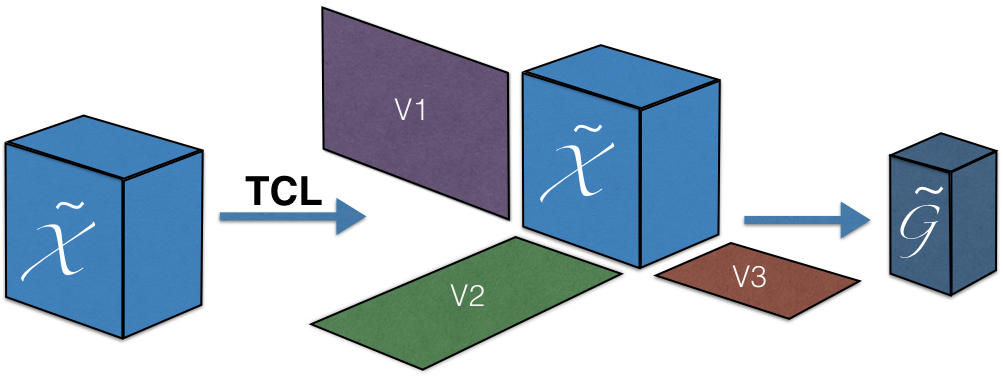}
\end{center}
\caption{A representation of the Tensor Contraction Layer (TCL) applied on a tensor of order 3.
The input tensor \(\mytensor{X}\) is contracted into a low-dimensionality core \(\mytensor{G}\).
}
\label{fig:TCL_visual}
\end{figure}

Given the enormous computational costs
for both training and running inference in 
these networks,
we desire techniques that preserve high accuracy 
while reducing the number of parameters 
in the network. 
Notable work in this direction includes approaches 
to induce and exploit sparsity in the parameters 
during training \cite{han2015deep}.

\section{Tensor Contraction Layer}
\label{seq:TCL}

In this paper, we propose to incorporate 
the tensor contraction 
into convolutional neural networks 
as an end-to-end trainable layer,
applying it to the third order activation tensor
output by the final convolutional layer.

In particular, given an activation tensor \(\mytensor{X}\) 
of size \( \left( D_1, \cdots, D_N \right) \), 
we seek a low dimensional core \(\mytensor{G}\) 
of smaller size \( \left( R_1, \cdots, R_N \right) \) such that: 
\begin{equation}
\mytensor{G} = 
\mytensor{X} \times_1 \mymatrix{V}^{(1)} 
		  \times_2  \mymatrix{V}^{(2)} \times
		  \cdots
          \times_N \mymatrix{V}^{(N)}
\end{equation}
with \(\mymatrix{V}^{(k)} \in \myR^{R_k, D_k}, k \in (1, \cdots, N)\).

We leverage this formulation and define a new layer 
that takes the activation tensor \(\mytensor{X}\) 
obtained from a previous layer 
and applies such a projection to it (Figure.~\ref{fig:TCL_visual}).
We optimize the projection factors \( \left(\mymatrix{V}^{(k)}\right)_{k \in [1, \cdots N]}\) 
to obtain a low dimensional projection 
of the activation tensor 
as the output of the layer. 
We learn the projection factors by backpropagation
jointly with the rest of the network's parameters. 
We call this new layer the tensor contraction layer 
and denote by \emph{size--\(\left(R_1, \cdots, R_N\right)\) TCL}, or \emph{TCL--\(\left(R_1, \cdots, R_N\right)\)}
a TCL producing a contracted output of size \(\left(R_1, \cdots, R_N\right)\).

The gradients with respect to each of the factors 
can be derived easily from \ref{eq:unfold_tucker}. 
Specifically, for each \(k \in {1, \cdots, N}\), 
we use the following equivalences:

\scriptsize
\begin{align*}
\myd{\mytensor{G}}{\mymatrix{V}^{(k)}}=&
\myd{\mytensor{X} \times_1 \mymatrix{V}^{(1)} 
		  \times_2  \mymatrix{V}^{(2)} \times
		  \cdots
          \times_N \mymatrix{V}^{(N)}}{\mymatrix{V}^{(k)}}
		  = \myd{\mytensor{G}_{[k]}}{\mymatrix{V}^{(k)}} \\
= & \myd{\mymatrix{V}^{(k)} \mymatrix{X}_{[k]}
    				   \left(\mymatrix{V}^{(1)}
                       \otimes \cdots 
                       \mymatrix{V}^{(k-1)} 
                       \otimes \mymatrix{V}^{(k+1)}
                       \otimes \cdots
                       \otimes \mymatrix{V}^{(N)} \right)^T
             }{\mymatrix{V}^{(k)}}\\
\label{eq:layer_gradient}
\end{align*}
\normalsize

\begin{figure}
\begin{center}
\includegraphics[width=7cm, height=10cm]{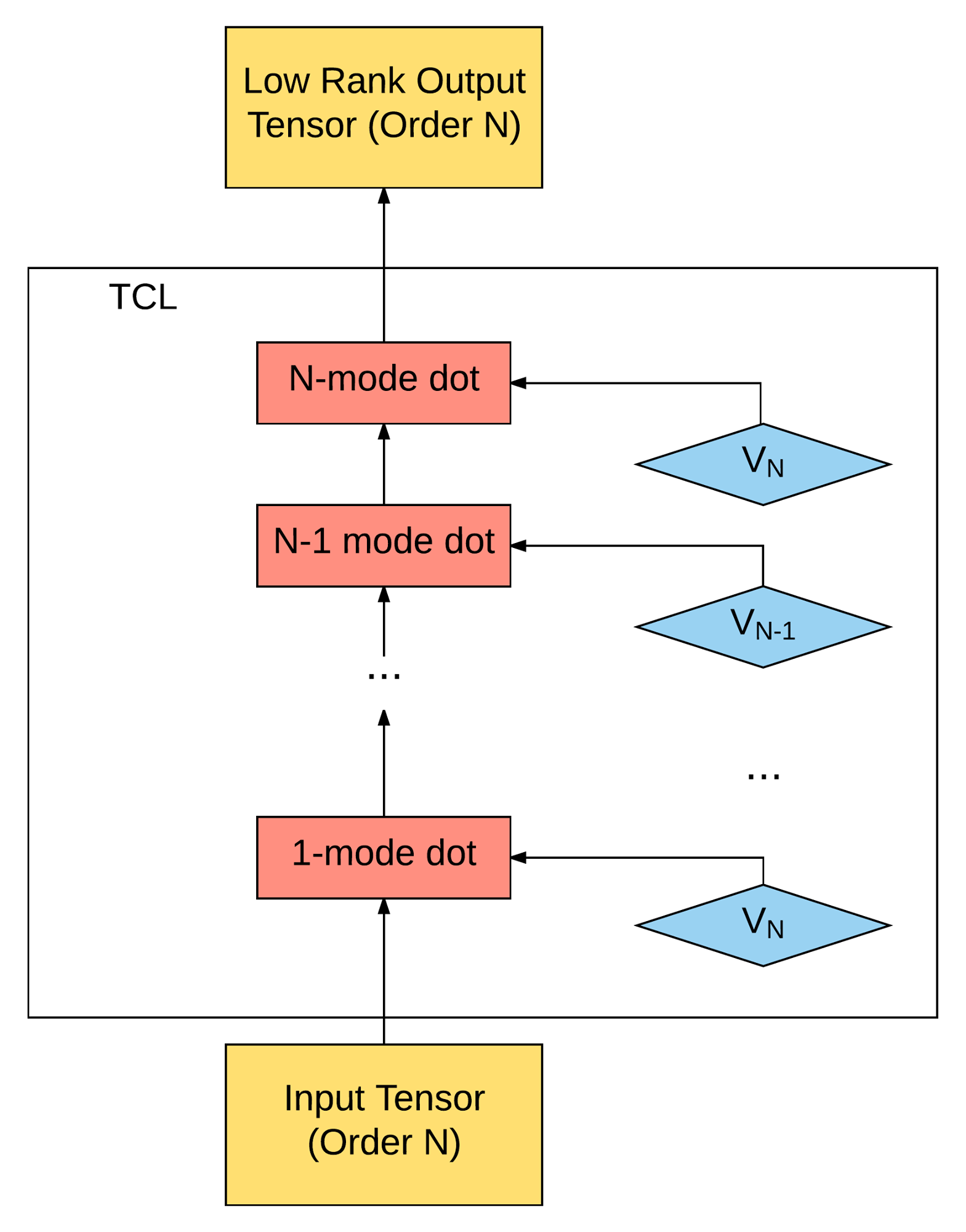}
\end{center}
\caption{A representation of the symbolic graph of the Tensor Contraction Layer.}
\label{fig:TCL_sym}
\end{figure}

In practice, with minibatch training, 
we might think of the first mode of an activation tensor 
as corresponding to the batch-size. 
Technically, it is possible to applying a transformation 
along this dimension too, 
but we leave this consideration for future work. 
It is trivial to address this case by either 
starting the \(n-\)mode products at the second mode 
or by setting the first factor 
to be the Identity and not optimize over it. 
Therefore, in the remainder of the paper,
we consider the activation tensor 
for a single sample for clarity, 
without loss of generality.

Figure.~\ref{fig:TCL_sym} presents the symbolic graph
of the tensor contraction layer. 
Note that when taking the \(n\)-mode product 
over different modes, 
the order in which the \(n\)-mode products are computed 
does not matter.

\subsection{Complexity of the TCL}

In this section, we detail the number of parameters 
and complexity of the tensor contraction layer.

\paragraph{Number of parameters}
Let \(\mytensor{X}\) be an activation tensor of size \( \left( D_1, \cdots, D_N \right) \) which we pass through a size--\( \left(R_1, \cdots, R_N \right) \) tensor contraction layer. 

This TCL has a total of \( \sum_{k=1}^{N} D_k \times R_k \) parameters 
(corresponding to the factors of the \(N\) \(n-\)mode products)
and produces as input a tensor of size \( \left(R_1, \cdots, R_N \right) \).

By comparison, a fully-connected layer producing an output of the same size,
i.e. with \(H = \prod_{k=1}^{N} R_k \) hidden units, 
and taking the same (flattened) tensor as input 
would have a total of \( \prod_{k=1}^{N} D_k \times \prod_{k=1}^{N} R_k \) parameters.  


\begin{table*}[ht!]
\begin{center}
	\small
	\rowcolors{2}{gray!25}{white}
	\begin{tabularx}{\linewidth}{ c | Y | c | c | Y | Y }
    	\rowcolor{gray!50}
		\textbf{Method} & \textbf{Added TCL} & \textbf{\(1^{st}\) fully-connected} & \textbf{\(2^{nd}\) fully-connected} &  \textbf{Accuracy (\%)} & \textbf{Space savings (\%)}\\
        \hline 
		Baseline &  - & 4096 hidden units & 4096 hidden units & 65.41 & 0\\
		\hline 
		Added TCL &  TCL--(256, 3, 3) & 4096 hidden units & 4096 hidden units & 65.53 & -0.25\\
        \hline
        Added TCL & TCL--(192, 3, 3) & 3072 hidden units & 3072 hidden units & 65.92 &  43.28 \\
		\hline
        Added TCL & TCL--(128, 3, 3) & 2048 hidden units & 2048 hidden units & \textbf{66.57} & 74.49 \\
        \hline
        1 TCL substitution & - & TCL--(256, 3, 3) & 4096 hidden units & 65.52 & 62.77 \\
        \hline
        1 TCL substitution & - & TCL--(192, 3, 3) & 3072 hidden units & 65.95 & 78.72 \\
        \hline
        1 TCL substitution & - & TCL--(128, 3, 3) & 2048 hidden units & 64.95 & 90.25\\
        \hline
        2 TCL substitutions & - & TCL--(256, 3, 3) & TCL--(256, 3, 3) & 62.98 & 98.64\\
        \hline
        2 TCL substitutions & - & TCL--(192, 3, 3) & TCL--(144, 3, 3) & 62.06 & \textbf{99.22}\\
	\end{tabularx}
    \normalsize
    \caption{Results with AlexNet on CIFAR100. The first column presents the method, the second specifies whether a tensor contraction was added and when this is the case, the size of the TCL. Columns 3 and 4 specify the number of hidden units in the fully-connected layers or the size of the TCL used instead when relevant. Column 5 presents the top-1 accuracy on the test set. Finally, the last column presents the reduction factor in the number of parameters in the fully-connected layers 
(which represent more than 80\% of the total number of parameters of the networks) 
where the reference is the original network without any modification (\emph{Baseline}).}
    \label{tab:alexnet_cifar}
\end{center}
\end{table*}

\paragraph{Complexity}
As previously exposed, one way to look at the TCL 
is as a series of matrix multiplications 
between the factors of the contraction 
and the unfolded activation tensor.
Let's place ourselves in the setting previously detailed
with an activation tensor \(\mytensor{X}\) of size \( \left( D_1, \cdots, D_N \right) \) 
and a TCL--\( \left(R_1, \cdots, R_N \right) \)
of complexity \(O(C_{\text{TCL}})\). 
We can write \(C_{\text{TCL}} = \sum_{k=1}^N C_k \) 
where \(C_k \) is the complexity of the \(k^{\text{th}}\) \(n-\)mode product. 
Note that the order in which the products are taken does not matter
due to the commutativity of the \(n-\)mode product over disjoint modes (e.g. it is commutative for \(\mytensor{X} \times_i \mymatrix{U}^{(i)} \times_j \mymatrix{U}^{(j)}\) as long as \( i \neq j\)).
However, for illustrative purposes, 
we consider them to be done in order, 
from the first mode to the \(N^{\text{th}}\). 
We then have:

\begin{equation}
	C_k = R_k \times D_k \prod_{i=1}^{k-1}R_i \prod_{j=k+1}^N D_j
\end{equation}
It follows that the overall complexity of the TCL is:
\begin{equation}
	C_{\text{TCL}} = \sum_{k=1}^N \prod_{i=1}^{k}R_i \prod_{j=k}^N D_j
\end{equation}

\paragraph{Comparison with a fully-connected layer}
A fully-connected layer with \(H\) hidden units has complexity \(O(C_{\text{FC}})\), with:
 \begin{equation}
    C_{\text{FC}} = H \prod_{i=1}^N D_i
 \end{equation}

Consider a TCL that maintains the size of its input, i.e., for any \(k\) in \(\myrange{1}{N}\), 
\(R_k = D_k\). 
In other words, \( C_k = D_k \prod_{i=1}^N D_i\). Therefore,
\begin{equation}
  C_{\text{TCL}} = \sum_{k=1}^N D_k \prod_{i=1}^N D_i
\end{equation}

By comparison, a fully-connected layer that also maintains the size of its input, i.e. \(H = \prod_{k=1}^N D_k\), would have a complexity of: 
 \begin{equation}
    C_{\text{FC}} = \left( \prod_{i=1}^N D_i \right)^2
 \end{equation}
 
Notice the product in the fully-connected case versus a sum for the TCL case.

\subsection{Incorporating TCL in a network}
We see several straightforward ways 
to incorporate the TCL
into existing neural network architectures.

\paragraph{TCL as An Additional Layer}
First, we can insert a tensor contraction layer 
following the last pooling layer, 
reducing the dimensionality of the activation tensor 
before feeding it to the subsequent two fully-connected layers 
and softmax output of the network.
In general, flattening induces a loss of information. 
By applying tensor contraction we reduce dimensionality 
efficiently by leveraging the multidimensional dependencies 
in the activation tensor.

\paragraph{TCL as Replacement of a Fully Connected Layer}
We can also incorporate the TCL into existing architectures
by completely replacing fully-connected layers.
This has the advantage of significantly reducing 
the number of parameters in our model.
Concretely, consider an activation tensor 
of size \(\left(256, 7, 7\right)\) 
that is fed to either a fully-connected layer 
(after having been flattened) or to a TCL. 
A fully-connected layer with \(4096\) hidden units 
has \(256 \times 7 \times 7 \times 4096 = 51,380,224\) parameters. 
A TCL that preserves the size of its input, on the other hand, 
only has \(256^2 + 7^2 + 7^2 = 1,712,622\) parameters.
The TCL has \(30\) times fewer parameters than the fully-connected layer. 
Similarly, a TCL--\( \left(128, 5, 5\right)\) 
(approximately half size) will have only
\( 256 \times 128 + 7 \times 5 + 7 \times 5 = 32,838\) parameters, 
or \(1,500\) times fewer parameters than a fully-connected layer.

\section{Experiments}
\label{seq:experiments}


\begin{table*}[ht!]
\begin{center}
	\small
	\rowcolors{2}{gray!25}{white}
	\begin{tabularx}{\linewidth}{ c | Y | c | c | Y | c}
    	\rowcolor{gray!50}
		\textbf{Method} & \textbf{Added TCL} & \textbf{\(1^{st}\) fully-connected} & \textbf{\(2^{nd}\) fully-connected} &  \textbf{Accuracy (\%)} & \textbf{Space savings (\%)} \\
        \hline 
		Baseline &  - & 4096 hidden units & 4096 hidden units & \textbf{69.98} & 0 \\
		\hline 
		Added TCL &  TCL--(512, 3, 3) & 4096 hidden units & 4096 hidden units & \textbf{70.07} & -0.73  \\
        \hline
        Added TCL & TCL--(384, 3, 3) & 3072 hidden units & 3072 hidden units & 68.56 & 42.99 \\
		\hline
        Added TCL & TCL--(256, 3, 3) & 2048 hidden units & 2048 hidden units & 67.57 & 74.35 \\
        \hline
        1 TCL substitution & - & TCL--(512, 3, 3) & 4096 hidden units & 69.71 & 45.8 \\
        \hline
        1 TCL substitution & - & TCL--(384, 3, 3) & 3072 hidden units & 68.83 & 69.16 \\
        \hline
        1 TCL substitution & - & TCL--(256, 3, 3) & 2048 hidden units & 68.51 & 85.98 \\
        \hline
        2 TCL substitutions & - & TCL--(512, 3, 3) & TCL--(512, 3, 3) & 67.20 & 97.27 \\
        \hline
        2 TCL substitutions & - & TCL--(384, 3, 3)  & TCL--(288, 3, 3) & 67.38 & \textbf{98.43} \\
	\end{tabularx}
    \normalsize
    \caption{Results obtained on CIFAR100 using a VGG-19 network architecture with different variations of the Tensor Contraction Layer. In all cases we report Top-1 Accuracy and space savings with respect to the baseline. As observed with the AlexNet, TCL allows for large space savings with minimal impact on performance and even improvement in some cases.}
    \label{tab:vgg_cifar}
\end{center}
\end{table*}

Our experiments investigate the representational power 
of the TCL, 
demonstrating results on the CIFAR100 dataset \cite{cifar}.
Subsequently, we offer some preliminary results on the ImageNet 1k dataset \cite{imagenet}. 
We hypothesize that a TCL 
can efficiently represent an activation tensor 
for processing by subsequent layers of the network, 
allowing for a large reduction in parameters without a reduction in accuracy. 

We conduct our investigation on CIFAR100 
using the AlexNet \cite{alexnet} and VGG \cite{vgg} architectures, 
each modified to take $32\times32$ images as inputs. 
We also present results with a traditional AlexNet on ImageNet. 
In all cases we report the accuracy (top-1) as well as the space saved, which we quantify as:
\[
\text{space savings} = 1 - \frac{n_{\text{TCL}}}{n_{\text{original}}}
\]
where \(n_{\text{original}}\) is the number of parameters in the fully-connected layers of the standard network and \(n_{\text{TCL}}\) is the number of parameters in the fully-connected layers of the network modified to include the TCL.

To avoid vanishing or exploding gradients, and to make the TCL more robust to changes in the initialization of the factors, we added a batch normalization layer \cite{ioffe2015batch} before and after the TCL.

\subsection{Results on CIFAR100}
The CIFAR100 dataset is composed of 100 classes 
containing 600 $32\times32$ images each, 
with 500 training images and 100 testing images per class. 
In all cases, we report performance on the testing set 
in term of accuracy (Top-1). 
We implemented all models using the MXNet library \cite{mxnet} 
and ran all experiments training with data parallelism across multiple GPUs on Amazon Web Services, 
with two NVIDIA k80 GPUs.

Because both the original AlexNet and VGG architectures 
were defined for the ImageNet data set, 
which has a larger input image size, 
to adapt them for CIFAR100 by adjusting the stride size 
on the input convolution layer of both networks 
so that they would take $32\times32$ input images. 
We investigate two sets of experiments, described below.
\begin{description}
\item[Added TCL]
In the first experiments, we added a TCL as additional layer 
after the last pooling layer 
and perform the contraction 
along the two spacial modes of the image, 
leaving the modes corresponding to the channel and the batch size untouched.
We gradually reduced the number of hidden units in these last two layers with and without the TCL included and retrain the nets until convergence to demonstrate how the TCL can learn more compact representations without compromising accuracy.

\item[TCL substitution] In this case, we completely replace one or both of the fully-connected layers 
by a tensor contraction layer. 
We reduce the number of hidden units in the subsequent layers proportionally to the reduction in the size of the activation tensor.
\end{description}

\paragraph{Network architectures}
We experimented with an AlexNet, 
with an adjusted stride and filter size 
in the final convolutional layer.
From the last convolutional layer,
we get an activation tensor of size 
\(\left(\textit{batch\_size}, 256, 3, 3\right)\).
Similarly, in the case of the VGG network, we obtain activation tensors of size
\(\left(\textit{batch\_size}, 512, 3, 3\right)\). 
We experiment with several variations 
of the tensor contraction layer.
First, we consider the case where we project the activations
to a tensor of identical shape.
Additionally, we evaluate the effect 
of reducing the dimensionality of the activation tensor 
by 25\% and by 50\%. 
For AlexNet, because the spatial modes 
already compact are already, 
we preserve the spatial dimensions, 
and reduce dimensionality along the channel.

\begin{table*}[ht!]
\begin{center}
\small
	\rowcolors{2}{gray!25}{white}
	\begin{tabularx}{\linewidth}{ c | Y | c | c | Y | Y }
    	\rowcolor{gray!50}
		\textbf{Method} & \textbf{Additional TCL} & \textbf{\(1^{st}\) fully-connected} & \textbf{\(2^{nd}\) fully-connected} &  \textbf{Accuracy (in \%)} & \textbf{Space savings (\%)}\\
        \hline 
		Baseline &  - & 4096 hidden units & 4096 hidden units & 56.29 & 0 \\
		\hline 
		Added TCL &  TCL--(256, 5, 5) & 4096 hidden units & 4096 hidden units & \textbf{57.54} & -0.11 \\
        \hline 
		Added TCL &  TCL--(200, 5, 5) & 3276 hidden units & 3276 hidden units & 56.11 & 35.36 \\
        \hline 
        TCL substitution & - & TCL--(256, 5, 5) & 4096 hidden units & 56.57 & \textbf{35.49} \\
	\end{tabularx}
    \normalsize
    \caption{Results obtained with AlexNet on ImageNet, for a standard AlexNet (\emph{baseline}), with an added Tensor Contraction Layer (\emph{Added TCL}) and by replacing the first fully-connected layer with a TCL (\emph{TCL substitution}). Simply adding the TCL results in a higher performance while having a minimal impact on the number of parameters in the fully connected layers. By reducing the size of the TCL or using a TCL to replace a fully connected layer, we can obtain a space savings of more than 35\% with virtually no deterioration in performance.}
    \label{tab:alexnet_imagenet}
\end{center}
\end{table*}

\subsubsection{Results}
Table~\ref{tab:alexnet_cifar} summarizes our results on CIFAR100 using the AlexNet, while results with VGG are presented in Table~\ref{tab:vgg_cifar}. 
The first column presents the method, 
the second specifies whether a tensor contraction 
was added and when this is the case, 
the size of the contracted core. 
Columns 3 and 4 specify the number of hidden units 
in the fully connected layers 
or the size of the TCL used instead when relevant. 
Column 5 presents the top-1 accuracy on the validation. Finally, the last column presents the reduction factor 
in the number of parameters in the fully connected layers (which represent, as previously mentioned, 
more than 80\% of the total number of parameters 
of the networks) where the reference 
is the original network without any modification (\emph{Baseline}).

A first observation is that adding a tensor contraction layer (\emph{Added TCL} in Tables ~\ref{tab:alexnet_cifar} and \ref{tab:vgg_cifar}) consistently increases performance while having minimal impact on the overall number of parameters. Replacing the first fully-connected layer (\emph{1 TCL substitution} in the Tables) allows us to reduce the number of parameters in the fully connected layers by a factor of more than $3$, while observing the same performance as the original network. By replacing both fully connected layers (\emph{2 TCL substitutions} in the Tables) we can obtain a reduction of more than $92$\(\times\), with only a $2.5\%$ decrease in performance.

\subsection{Results on ImageNet}

In this section, we present preliminary experiments using the larger ILSVRC 2012 (ImageNet) dataset \cite{imagenet}, using the AlexNet architecture. 
ImageNet is composed of $1.2$ millions image 
for testing and 50,000 for validation 
and comprises 1,000 labeled classes. 

For these experiments, we trained each network 
simultaneously on 4 NVIDIA k80 GPUs 
using data parallelism and report preliminary results.
We report Top-1 accuracy on the validation set, 
across all 1000 classes. All experiments were run using the same setting. 

\paragraph{Network architecture}
We use a standard AlexNet \cite{alexnet}. From the last convolutional layer, we get an activation tensor of size \(\left(\textbf{batch\_size}, 256, 5, 5\right)\). 
As in the CIFAR100 case, we experiment with several variations of the tensor contraction layer. 
We first insert a TCL before the fully-connected layers, 
either a size-preserving TCL (i.e. projecting to a tensor of the same size) 
or with a smaller size TCL and a proportionally smaller number 
of hidden units in the subsequent fully-connected layers. 
We then experiment with replacing completely the first fully-connected layer with a TCL. 

\subsubsection{Results}
In Table \ref{tab:alexnet_imagenet}
we summarize the results from a standard AlexNet (\emph{Baseline}, first row), with an added tensor contraction layer (\emph{Added TCL}) that preserves the dimensionality of its input (row 2) or reduces it (last row). 
We also report result for substituting 
the first fully connected layer 
with a TCL (\emph{1 TCL substitution}, last row).
Simply adding the TCL improves performance 
while the increase in number of parameters 
in the fullly connected layers is negligible.
We can obtain similar performance by first adding a TCL to reduce the dimensionality of the activation tensor and reducing the number of hidden units in the fully-connected layers, leading to a large space saving with virtually no decrease in performance. Replacing the first fully-connected layer with a size-preserving TCL results in a similar space savings while maintaining the same performance as the standard network.

\section{Discussion}
We introduced a new neural network layer 
that performs a tensor contraction 
on an activation tensor to yield 
a low dimensional representation of it. 
By exploiting the natural multi-linear structure 
of the data in the activation tensor, 
where each mode corresponds to a distinct modality 
(i.e. the dimensions of the image and the channels), 
we are able to decrease the size 
of the data representation 
passed to subsequent layers in the network 
without compromising accuracy on image recognition tasks.

The biggest practical contribution of the TCL is the drastic reduction in the number of parameters with little to no performance penalty. This also allows neural networks to perform faster inference with fewer parameters by increasing their representational power. 
We demonstrated this via the performance of TCLs 
on the widely used CIFAR100 dataset 
with two established architectures, namely AlexNet and VGG. 
We also show results with AlexNet on the ImageNet dataset. 
Our proposed tensor contraction layer seems to be able to capture the underlying structure in the activation tensor and improve performance when added to an existing network. When we replace fully-connected layers with TCLs, 
we significantly reduce the number of parameters 
and nevertheless maintain (or in some cases even improve) performance.

Going forward, we plan to extend our work 
to more network architectures, 
especially in settings where raw data or learned representations exhibit natural multi-modal structure 
that we might capture via high-order tensors. 
We also endeavor to advance our experimental study
of TCLS for large-scale, high-resolutions vision datasets.  
Given the time required to train a 
large network on such datasets 
we are investigating ways to reduce the dimension 
of the tensor contractions of an already trained model and simply fine tune. 
In addition, recent work \cite{shi2016tensor} has shown that
new extended BLAS primitives can avoid transpositions needed to compute the tensor contractions.
This will further speed up the computations and we plan to implement it in future.
Furthermore, we will look into methods 
to induce and exploit sparsity in the TCL, 
to understand the parameter reductions this method can yield 
over existing state-of-the-art pruning methods. 
Finally, we are working on an extension to the TCL: 
a tensor regression layer to replace both 
the fully-connected and final output layers, 
potentially yielding increased accuracy with even greater parameter reductions. 

{\small
\bibliographystyle{ieee}
\bibliography{refs}
}

\end{document}